\documentclass[10pt,twocolumn]{article}
\usepackage[a4paper,margin=0.75in,columnsep=0.22in]{geometry}
\usepackage[T1]{fontenc}
\usepackage{microtype}
\usepackage{graphicx,booktabs,amsmath,amssymb,multirow,xcolor}
\usepackage{tikz}
\usetikzlibrary{positioning,arrows.meta,decorations.pathreplacing}
\usepackage[round,authoryear]{natbib}
\usepackage[hidelinks]{hyperref}
\usepackage{authblk}

\title{MRI Super-Resolution with RCDM/WaveMix\\[2pt] and Task-Aware Segmentation}
\author[1]{Kavitha Viswanathan}
\author[2]{Harsh Choudhary}
\author[2]{Amit Sethi}
\affil[1,2]{Department of Electrical Engineering, Indian Institute of Technology Bombay, Mumbai 400\,076, India}
\date{}

\begin{document}
\twocolumn[
\begin{@twocolumnfalse}
\maketitle
\begin{abstract}
Super-resolution and quality enhancement of 1.5\,T brain MRI are normally
validated with image-fidelity metrics, although their purpose is to improve
downstream analysis. We study whether enhancement improves tissue
segmentation, and for which segmenters. We propose an unpaired, physics-guided
training pipeline for a lightweight ($\le$2.5\,M parameter) recurrent
convolutional enhancer: a six-module stochastic 1.5\,T degradation operator, a
residual adversarial network that adds scanner-specific texture without moving
anatomy, and a cycle-consistent objective with an anti-identity penalty that
rules out the copy solution. We then train U-Net, Swin-UNet and wavelet
token-mixing segmenters \citep{jeevan2023wavemix} from scratch on either raw or enhanced 1.5\,T images
of the same subjects, using identical labels and subject-level splits, for
three enhancer variants and two datasets. On ABIDE (41 held-out subjects,
FreeSurfer labels) enhancement significantly improves the wavelet segmenter
(mean Dice $+0.014$, Wilcoxon $p=3.5\times10^{-5}$; CSF $+0.018$, grey matter
$+0.013$), significantly degrades the U-Net ($-0.008$, $p=5.1\times10^{-4}$) and
leaves Swin-UNet unchanged. On IXI, whose labels come from FSL-FAST,
enhancement lowers Dice for all nine pairings, almost entirely through CSF;
we trace this to spatially implausible CSF voxels in the labels that penalise
smoother predictions. Enhancement of low-field MRI should therefore be
validated per downstream model and against reliable labels.
\end{abstract}
\bigskip
\end{@twocolumnfalse}]

\section{Introduction}\label{sec:intro}
Magnetic field strength largely determines the quality of structural brain MRI.
At 3\,T the signal-to-noise ratio is roughly twice that at 1.5\,T, tissue
contrast is higher and partial-volume blurring at tissue boundaries is lower.
Yet a large fraction of clinical, longitudinal and multi-site research data
has been, and continues to be, acquired at 1.5\,T. If 1.5\,T images could be
mapped to 3\,T appearance, analysis tools developed and tuned on 3\,T data
could be applied to them more reliably. This motivates learned enhancement and
super-resolution (SR) of brain MRI
\citep{pham2017brainsr,chen2018mdcsrn,zhao2021smore} and image-quality
transfer across field strengths
\citep{alexander2017iqt,lin2023lowfieldiqt,iglesias2023synthsr}.

Most such methods are validated with PSNR, SSIM or perceptual similarity to a
reference. These metrics answer whether the output looks like the target, not
whether it helps the analysis the target is used for. The distinction matters
for two reasons. First, a learned model can raise fidelity while inventing or
removing small structures; global metrics barely register this, but a
segmentation model will. Second, even an image that is objectively closer to
3\,T need not help every downstream model equally, because each model relies on
different image statistics \citep{maierhein2024metrics}. A fidelity gain is
therefore at best a proxy, and its relationship to the downstream task has to
be measured.

We measure it for brain tissue segmentation. The question we ask is concrete:
\emph{if a segmenter is trained and tested on enhanced 1.5\,T images instead of
the raw 1.5\,T images of the same subjects, with the same labels and split,
does Dice improve; and does the answer depend on the segmenter and on the label
source?} Training separate segmenters for each input type, rather than
applying a single frozen model, isolates the information content of the
enhanced images from the domain shift a frozen model would suffer.

\noindent Our contributions are:
\begin{enumerate}
\item A physics-guided, unpaired training pipeline for 1.5\,T enhancement that
combines a six-module stochastic degradation operator, a residual
adversarial texture model that cannot move anatomy, and a cycle-consistent
objective with an explicit anti-identity penalty (Section~\ref{sec:method}).
\item A lightweight recurrent convolutional enhancer, adapted from video
super-resolution, that treats adjacent slices as frames and has at most
2.5\,M parameters (Section~\ref{sec:sr}).
\item A controlled segmentation-based evaluation: three enhancer variants
crossed with three segmentation backbones, two datasets, subject-level
splits, paired non-parametric tests and multiple-comparison correction
(Sections~\ref{sec:protocol}--\ref{sec:results}).
\item Two findings with practical consequences: the effect of enhancement is
backbone-dependent, positive for a wavelet token-mixing segmenter and
negative for a U-Net; and automated labels can reverse its sign
(Sections~\ref{sec:results}--\ref{sec:discussion}).
\end{enumerate}

\section{Related work}\label{sec:related}
\textbf{Brain MRI super-resolution.} Early deep SR for brain MRI used 3D
convolutional networks trained on synthetically downsampled high-resolution
volumes \citep{pham2017brainsr}; adversarial training and densely connected
3D networks followed \citep{chen2018mdcsrn}. Self-supervised approaches such as
SMORE \citep{zhao2021smore} exploit the higher in-plane resolution of
anisotropic acquisitions and avoid external training pairs. All of these
report fidelity metrics; downstream effects, when studied, are usually
assessed with a single analysis tool.

\textbf{Image-quality transfer across field strengths.} Image-quality transfer
learns a mapping from low- to high-quality acquisitions
\citep{alexander2017iqt}, and has been extended to very-low-field MRI with a
stochastic degradation model that simulates low-field images from
high-field ones \citep{lin2023lowfieldiqt}. SynthSR \citep{iglesias2023synthsr}
and SynthSeg \citep{billot2023synthseg} take a complementary route, training on
synthetic images generated from label maps so that models become robust to
contrast and resolution. Our degradation operator is in the same spirit as
the stochastic simulation of \citet{lin2023lowfieldiqt}, but targets the
1.5\,T to 3\,T gap and is combined with a learned residual texture model.

\textbf{Unpaired and cycle-consistent training.} Cycle consistency
\citep{zhu2017cyclegan} is the usual tool when paired data are unavailable.
A known failure mode is a near-identity mapping that satisfies the cycle
constraint trivially. We address it with an explicit penalty on outputs that
are not sharper than their input.

\textbf{Task-based evaluation.} The recommendation that validation metrics be
chosen according to the downstream use of an algorithm is now established
\citep{maierhein2024metrics}. We apply it to enhancement, and show that the
conclusion depends on the downstream model and the label source.

\section{Method}\label{sec:method}
Let $\mathbf{x}\in\mathbb{R}^{S\times H\times W}$ be a stack of $S$ adjacent
axial 3\,T slices, and $\mathbf{c}(\mathbf{x})$ its centre slice. The goal is an
enhancer $G$ that maps a stack of 1.5\,T slices to a 3\,T-like centre slice of
the same size.

\subsection{Physics-inspired stochastic degradation}\label{sec:degr}
The degradation operator $\mathcal{D}$ composes six modules, each applied with
probability $p_i$ and with parameters drawn uniformly from the given ranges
at every call (Table~\ref{tab:degr}). Each module models one physical
difference between 1.5\,T and 3\,T acquisitions:
\begin{enumerate}
\item \emph{Resolution loss}: anisotropic Gaussian blur,
$\mathbf{x}\leftarrow\mathbf{x}*\mathcal{G}_{\sigma_{xy},\sigma_z}$.
\item \emph{Lower SNR}: complex Gaussian noise added in $k$-space followed by
magnitude reconstruction,
$\mathbf{x}\leftarrow\lvert\mathcal{F}^{-1}(\mathcal{F}\mathbf{x}+\eta)\rvert$,
$\eta\sim\mathcal{CN}(0,s^{2})$, which produces the Rician statistics of
magnitude images \citep{gudbjartsson1995rician}.
\item \emph{Receive-field inhomogeneity}: a smooth multiplicative bias field
obtained by bicubic upsampling of a coarse random grid.
\item \emph{Contrast difference}: a monotone intensity remap
$\mathbf{x}\leftarrow a\,\mathbf{x}^{\gamma}$ that shifts grey--white
contrast towards 1.5\,T statistics.
\item \emph{Residual noise}: image-domain Gaussian noise.
\item \emph{Partial volume}: bicubic down-sampling by a factor $f$ followed by
bicubic up-sampling to the original grid.
\end{enumerate}

\begin{table}[t]
\caption{Stochastic 1.5\,T degradation operator. Each module fires with
probability $p$; parameters are drawn uniformly from the ranges.}\label{tab:degr}
\centering\small
\begin{tabular}{@{}llc@{}}
\toprule
Module & Parameters & $p$\\
\midrule
Anisotropic blur & $\sigma_{xy}\!\in\![0.6,1.8]$, $\sigma_z\!\in\![0.8,2.2]$ & 0.95\\
$k$-space (Rician) noise & $s\in[0.02,0.08]$ & 0.90\\
Bias field & strength $\in[0.05,0.20]$ & 0.70\\
Intensity remap & $\gamma\in[0.80,1.25]$, $a\in[0.85,1.15]$ & 0.50\\
Image-domain noise & $\sigma\in[0.005,0.025]$ & 0.30\\
Partial-volume resampling & $f\in[1.3,1.8]$ & 0.40\\
\bottomrule
\end{tabular}
\end{table}

\subsection{Residual adversarial texture model}\label{sec:gan}
An analytic operator cannot reproduce site- and scanner-specific texture. We
therefore learn a residual on top of it,
\begin{equation}
\tilde{\mathbf{x}}=\mathrm{clip}\big(\mathcal{D}(\mathbf{x})+R(\mathcal{D}(\mathbf{x})),\,0,\,1\big),
\end{equation}
where $R$ is an encoder--decoder with six residual blocks, instance
normalisation and reflection padding ($\approx$8.4\,M parameters). Because $R$
only adds a residual to an image whose geometry is fixed by $\mathcal{D}$, it
can alter texture and contrast but has little freedom to move anatomy. A
two-scale PatchGAN discriminator \citep{wang2018patchgan} compares
$\tilde{\mathbf{x}}$ with real 1.5\,T slices using the least-squares
adversarial loss \citep{mao2017lsgan} plus feature matching, with learning
rates $2\times10^{-4}$ for $R$ and $10^{-4}$ for the discriminator. Seeded
stochastic passes of $\mathcal{D}$ followed by $R$ yield several distinct
1.5\,T-like versions of each 3\,T volume.

\subsection{Enhancer architecture}\label{sec:sr}
The enhancer is a lightweight recurrent convolutional network originally
designed for video SR \citep{viswanathan2025lowresource}. It combines (i)~a
single-tensor residual memory that carries information across the stack,
(ii)~a 2D Haar wavelet branch on the centre slice that conditions the fusion
blocks on its sub-bands, and (iii)~a 3D deformable convolution
\citep{zhu2019dcnv2} that aligns neighbouring slices implicitly. For volumes,
the ``frames'' of the group-of-frames (GOF) window are adjacent axial slices,
so temporal aggregation becomes through-plane aggregation, and the output has
the input resolution. We use three variants: GOF\,=\,3 and GOF\,=\,5 versions
($\approx$2.3\,M parameters each), and \emph{WaveMix-SR}, which adds a small
wavelet token-mixing \citep{jeevan2023wavemix} residual refiner (32 features,
4 blocks, residual scale 0.05; $\le$2.5\,M parameters in total).

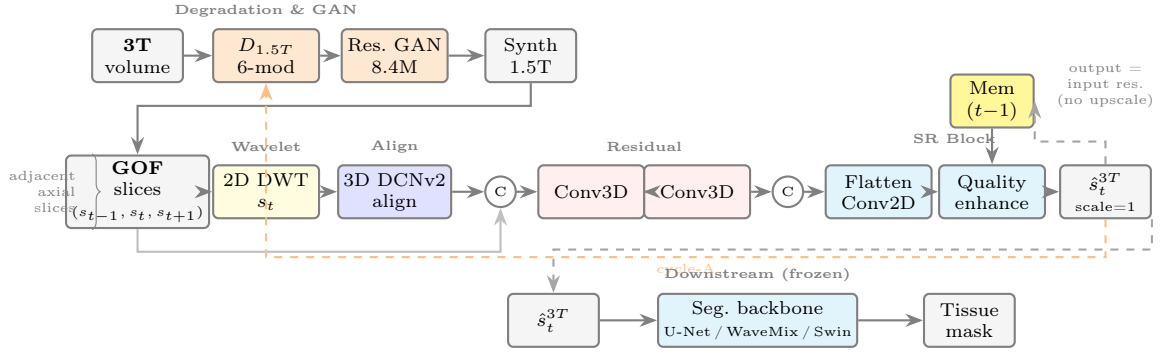
\begin{figure*}[!t]
  \centering
  \begin{tikzpicture}[
  font=\scriptsize,
  >=Stealth,
  every node/.style={inner sep=2pt},
  blk/.style={draw=black!50, thick, rounded corners=2pt,
              align=center, minimum height=7mm, minimum width=14mm},
  wavelet/.style={blk, fill=yellow!15},
  resid/.style={blk, fill=pink!25},
  aln/.style={blk, fill=blue!12},
  srblk/.style={blk, fill=cyan!10},
  memblk/.style={blk, fill=yellow!50, minimum width=11mm},
  ganblk/.style={blk, fill=orange!18},
  io/.style={blk, fill=gray!8, minimum width=12mm},
  cat/.style={circle, draw=black!50, fill=white, thick,
              minimum size=3.5mm, font=\tiny},
  arr/.style={->, thick, black!50},
  darr/.style={->, thick, black!35, dashed},
  lbl/.style={font=\tiny\bfseries, black!50},
]

\node[io]     (in3t) at (0,0)     {\textbf{3T}\\volume};
\node[ganblk] (deg)  at (1.7,0)   {$D_{1.5T}$\\6-mod};
\node[ganblk] (gan)  at (3.4,0)   {Res.\ GAN\\8.4M};
\node[io]     (s15)  at (5.2,0)   {Synth\\1.5T};
\draw[arr] (in3t)--(deg); \draw[arr] (deg)--(gan); \draw[arr] (gan)--(s15);
\node[lbl] at (1.7,0.6) {Degradation \& GAN};

\node[io]      (slc) at (0,-1.8)  {\textbf{GOF}\\slices\\{\tiny($s_{t-1},s_t,s_{t+1}$)}};
\node[wavelet] (dwt) at (1.7,-1.8){2D DWT\\$s_t$};
\node[aln]     (dcn) at (3.4,-1.8){3D DCNv2\\align};
\node[cat]     (c1)  at (4.8,-1.8){C};
\node[resid]   (r1)  at (6.0,-1.8){Conv3D};
\node[resid]   (r2)  at (7.4,-1.8){Conv3D};
\node[cat]     (c2)  at (8.6,-1.8){C};

\draw[arr] (slc)--(dwt); \draw[arr] (dwt)--(dcn);
\draw[arr] (dcn)--(c1);
\draw[arr] (c1)--(r1); \draw[arr] (r1)--(r2); \draw[arr] (r2)--(c2);
\draw[arr,black!25] (slc.south)--++(0,-3mm)-|(c1.south);
\node[lbl] at (1.7,-1.2) {Wavelet};
\node[lbl] at (3.4,-1.2) {Align};
\node[lbl] at (6.7,-1.2) {Residual};

\node[srblk]  (flat) at (9.8,-1.8){Flatten\\Conv2D};
\node[srblk]  (qe)   at (11.3,-1.8){Quality\\enhance};
\node[memblk] (mem)  at (11.3,-0.6){Mem\\$(t{-}1)$};
\node[io]     (out)  at (12.8,-1.8){\textbf{$\hat{s}_t^{3T}$}\\{\tiny scale=1}};
\draw[arr] (c2)--(flat); \draw[arr] (flat)--(qe); \draw[arr] (qe)--(out);
\draw[arr] (mem.south)--(qe.north);
\draw[darr] (out.north)--++(0,3mm)-|(mem.east);
\node[lbl] at (10.8,-1.1) {SR Block};

\draw[arr] (s15.south)--++(0,-3mm)-|(slc.north);
\draw[darr, orange!50] (out.south)--++(0,-5mm)--++(-11.1,0)
  node[midway,below,font=\tiny]{cycle-A}-|(deg.south);

\node[io]    (si)  at (5.5,-3.5) {$\hat{s}_t^{3T}$};
\node[srblk, minimum width=24mm] (seg) at (8.2,-3.5)
  {Seg.\ backbone\\{\tiny U-Net\,/\,WaveMix\,/\,Swin}};
\node[io]    (so)  at (11.0,-3.5){Tissue\\mask};
\draw[arr] (si)--(seg); \draw[arr] (seg)--(so);
\draw[darr] (out.south east)--++(0,-4mm)-|(si.north);
\node[lbl] at (8.2,-2.9) {Downstream (frozen)};

\draw[decorate, decoration={brace, amplitude=3pt, mirror}, gray]
  (-0.6,-2.3) -- (-0.6,-1.3)
  node[midway, left=4pt, font=\tiny, gray, align=right] {adjacent\\axial\\slices};
\node[font=\tiny, gray, align=center] at (12.8,-0.4)
  {output =\\input res.\\(no upscale)};

\end{tikzpicture}
  \caption{RCDM enhancer adapted for structural brain MRI. \textbf{Top:}
  a six-module stochastic degradation operator $D_{1.5T}$ and a residual GAN
  ($\sim$8.4\,M params) synthesise 1.5\,T-like slices from 3\,T volumes.
  \textbf{Middle:} a group of $S$ adjacent axial slices (GOF\,=\,3 or 5) enters
  the backbone \citep{viswanathan2025lowresource}: 2D Haar DWT of the centre
  slice (beige), 3D deformable alignment (blue), Conv3D residual blocks (pink),
  then flatten, Conv2D and quality-enhancement head (cyan) gated by a memory
  tensor (yellow). The output has the same resolution as the input
  (scale\,=\,1: quality enhancement, not spatial upscaling). Dashed orange:
  cycle-A re-degradation consistency. \textbf{Bottom:} a frozen 3\,T-trained
  segmentation backbone provides the downstream Dice evaluation.}
  \label{fig:arch}
\end{figure*}

\subsection{Training objective}\label{sec:loss}
The reconstruction loss between a prediction $\hat{y}$ and target $y$ is
\begin{multline}
\mathcal{L}_{\mathrm{rec}}(\hat{y},y)=\lVert\hat{y}-y\rVert_1
+0.2\,\big(1-\mathrm{SSIM}(\hat{y},y)\big)\\
+0.05\,\ell_{\mathrm{edge}}+0.03\,\ell_{\mathrm{Lap}}
+0.01\,\ell_{\mathrm{freq}}+0.01\,\ell_{\mathrm{con}},
\end{multline}
where the last four terms compare image gradients, Laplacians, Fourier
magnitudes and local contrast. The full objective is
\begin{equation}
\mathcal{L}=\mathcal{L}_{\mathrm{rec}}^{(1)}+0.5\,\mathcal{L}_{\mathrm{rec}}^{(2)}
+0.5\,\mathcal{L}_{\mathrm{cyc}}+0.1\,\mathcal{L}_{\mathrm{anti}}+0.1\,\mathcal{L}_{\mathrm{id}}.
\label{eq:total}
\end{equation}
\emph{Paired terms.} $\mathcal{L}_{\mathrm{rec}}^{(k)}=\mathcal{L}_{\mathrm{rec}}(G(\mathcal{D}_k(\mathbf{x})),\mathbf{c}(\mathbf{x}))$
for two independent random draws $\mathcal{D}_1,\mathcal{D}_2$ of the
operator, which exposes the enhancer to two degradations of the same anatomy
per step.

\emph{Cycle term.} For a GAN-generated 1.5\,T stack $\tilde{\mathbf{x}}$, the
enhanced output is re-degraded and compared with its input,
\begin{equation}
\mathcal{L}_{\mathrm{cyc}}=\big\lVert\mathbf{c}(\mathcal{D}(G(\tilde{\mathbf{x}})))-\mathbf{c}(\tilde{\mathbf{x}})\big\rVert_1
+0.2\big(1-\mathrm{SSIM}(\cdot,\cdot)\big).
\end{equation}
It constrains the enhancer on realistic 1.5\,T appearance for which no 3\,T
target exists.

\emph{Anti-identity term.} $\mathcal{L}_{\mathrm{cyc}}$ is minimised trivially
by an enhancer that returns a slightly smoothed copy of its input. With
$\mathrm{TV}(u)$ the mean absolute horizontal plus vertical finite difference,
\begin{equation}
\mathcal{L}_{\mathrm{anti}}=\max\!\big(0,\;\mathrm{TV}(\mathbf{c}(\tilde{\mathbf{x}}))
-0.95\,\mathrm{TV}(G(\tilde{\mathbf{x}}))\big)
\end{equation}
penalises any output that is not at least nearly as sharp as its input; the
factor 0.95 leaves room for removing genuine noise.

\emph{Identity term.} $\mathcal{L}_{\mathrm{id}}=\lVert G(\mathbf{x})-\mathbf{c}(\mathbf{x})\rVert_1$
on clean 3\,T input discourages changes to images that are already of high
quality.

The cycle and anti-identity terms are enabled after 3 epochs and the identity
term after 5, so that the enhancer first learns the paired mapping.

\section{Evaluation protocol}\label{sec:protocol}
\subsection{Data}
\textbf{ABIDE} \citep{martino2014abide}: multi-site 1.5\,T T1-weighted
volumes with FreeSurfer-derived \citep{fischl2012freesurfer} tissue labels.
\textbf{IXI} \citep{ixi2025}: 1.5\,T T1-weighted volumes whose tissue labels
are produced by FSL-FAST \citep{zhang2001fast}. All volumes are
skull-stripped, bias-field corrected, resampled to $1\,\mathrm{mm}^3$ and
intensity-normalised; labels are mapped to background, CSF, grey matter (GM)
and white matter (WM).

\subsection{Segmentation experiment}
For each dataset we create one subject-level split (30\% test; 15\% of the
remainder for validation; seed 42), stored and reused for every experiment, so
no subject contributes slices to more than one partition. For ABIDE the test
set contains 41 subjects. For each backbone and each enhancer variant we train
two segmenters from scratch with identical hyper-parameters: \textsc{raw},
trained and tested on real 1.5\,T slices, and \textsc{sr}, trained and tested on
the enhanced slices of the same subjects, with the same labels. The difference
between the two isolates the effect of enhancement on what a segmenter can
learn from the images.

The backbones are a U-Net \citep{ronneberger2015unet} ($\approx$0.5\,M
parameters), a Swin-UNet \citep{cao2022swinunet} ($\approx$1.7\,M) and a
wavelet token-mixing segmenter based on WaveMix \citep{jeevan2023wavemix}
($\approx$0.7\,M), chosen to represent convolutional, windowed-attention and
wavelet-mixing inductive biases at comparable, small size. As a sanity check
on the last, a seven-block configuration of the same segmenter reaches mIoU
0.8156 on a 15-class Cityscapes-style benchmark.

Training uses 2D axial slices (every fifth slice) resized to
$128\times128$, class-weighted cross-entropy (0.05/0.35/0.30/0.30 for
background/CSF/GM/WM) plus 0.5$\times$soft Dice loss, AdamW (learning rate
$10^{-3}$, cosine annealing to $10^{-5}$, weight decay $10^{-4}$), batch 8,
up to 50 epochs with early stopping (patience 12) and gradient clipping at 5.

\subsection{Statistics}
Dice and IoU are computed per subject over its test slices. \textsc{raw} and
\textsc{sr} are compared with the two-sided paired Wilcoxon signed-rank test
\citep{wilcoxon1945}; we also report the number of subjects improved. For the
12 backbone$\times$metric tests of the main comparison we apply Bonferroni
correction ($\alpha=0.05/12$).

\subsection{Enhancer training details}
The enhancer is trained with AdamW (cosine schedule over 100 epochs, best validation checkpoint retained; the logged WaveMix-SR run completed 21 epochs; initial learning rate
$10^{-4}$, cosine annealing), batch size 16 and 8 randomly positioned stacks per
volume, seed 42. Fidelity is measured on synthetic 1.5\,T/3\,T pairs from held-out
volumes with PSNR, SSIM and MicroSSIM
\citep{ashesh2024microssim}, which weights small high-contrast structures more
heavily than global SSIM.

\section{Results}\label{sec:results}
\subsection{Reconstruction fidelity}
\begin{figure*}[t]
\centering
\includegraphics[width=0.9\textwidth]{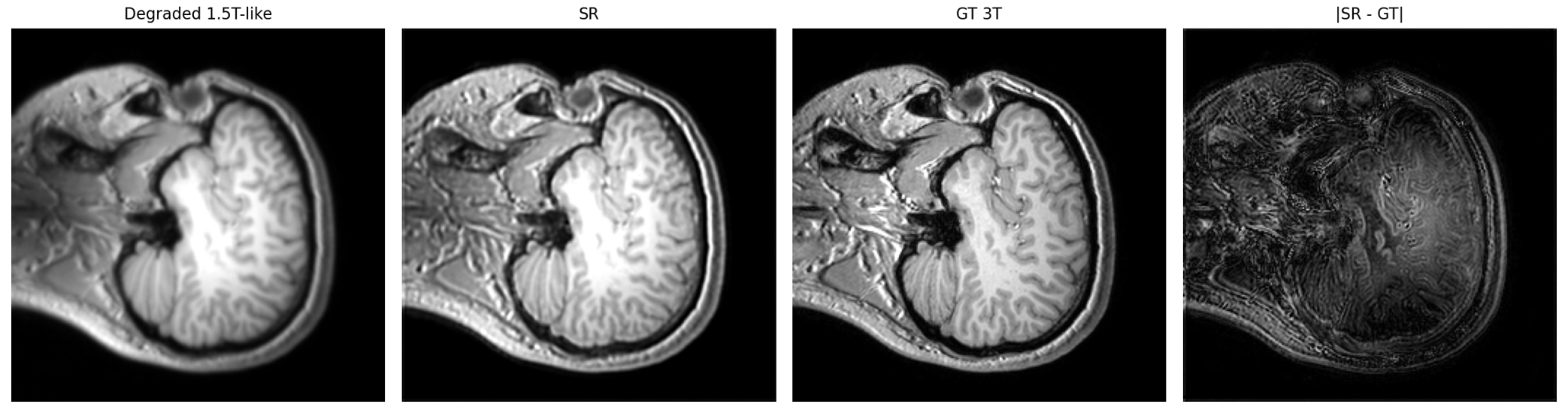}
\caption{Synthetic 1.5\,T input, enhanced output, 3\,T target and absolute
error for a held-out slice.}\label{fig:recon}
\end{figure*}

Table~\ref{tab:fid} reports validation fidelity on held-out synthetic
1.5\,T/3\,T pairs for the WaveMix-SR run trained on IXI, the only run for which
the per-epoch validation log is available. Relative to its degraded input the
enhancer gains about 1.4\,dB PSNR, 0.086 SSIM and 0.083 MicroSSIM at the best
epoch. The absolute values are low because the degradation operator is
severe by design and because PSNR is computed over whole slices including
background. Fig.~\ref{fig:recon} shows an example; the residual error
concentrates at the cortical ribbon and at vessel and skull boundaries.
\begin{table}[t]
\caption{Validation fidelity on synthetic 1.5\,T/3\,T pairs (IXI). Best epoch by SSIM.}
\label{tab:fid}
\centering\small
\begin{tabular}{@{}lccc@{}}
\toprule
 & PSNR$\uparrow$ & SSIM$\uparrow$ & MicroSSIM$\uparrow$\\
\midrule
Input (epoch 6) & 22.37 & 0.790 & 0.775\\
Best (epoch 6) & \textbf{23.85} & \textbf{0.876} & \textbf{0.859}\\
Final (epoch 21) & 23.71 & 0.861 & 0.844\\
\bottomrule
\end{tabular}
\end{table}

\subsection{Segmentation on ABIDE}
\begin{figure}[t]
\centering
\includegraphics[width=\columnwidth]{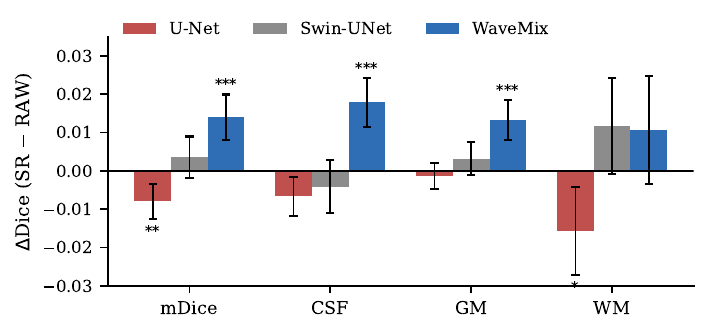}
\caption{ABIDE paired $\Delta$Dice (SR$-$RAW), mean $\pm$95\% CI over 41
subjects. Stars: Bonferroni-corrected Wilcoxon $p$
(* $<$0.05, ** $<$0.01, *** $<$0.001).}\label{fig:delta}
\end{figure}

Table~\ref{tab:abide} gives the main comparison for the GOF\,=\,5 enhancer and
Fig.~\ref{fig:delta} the paired differences. Three outcomes appear on the same
subjects and labels. The wavelet segmenter improves in mean Dice
($+0.0139$, $p=3.5\times10^{-5}$, 30/41 subjects), CSF ($+0.0178$,
$p=7.8\times10^{-7}$, 36/41) and GM ($+0.0133$, $p=3.2\times10^{-6}$, 36/41),
The U-Net loses mean
Dice ($-0.0079$, $p=5.1\times10^{-4}$, only 9/41 improved), mostly through
WM ($-0.0157$, $p=2.0\times10^{-3}$). Swin-UNet changes by less than 0.012 on every class
and no change is significant. After Bonferroni correction the wavelet-segmenter
gains in mDice, CSF and GM and the U-Net losses in mDice and WM remain
significant.

\begin{table}[!htbp]
\caption{ABIDE segmentation (GOF\,=\,5 enhancer, $n{=}41$).
$\Delta$: mean paired difference; $k$: subjects improved;
$p$: Wilcoxon. Bold: significant after Bonferroni ($\times12$).}
\label{tab:abide}
\centering\scriptsize\setlength{\tabcolsep}{2pt}
\begin{tabular}{@{}llcccr@{}}
\toprule
 & Metric & RAW & SR & $\Delta$ ($k$) & $p$\\
\midrule
\multirow{4}{*}{\rotatebox{90}{U-Net}}
 & mDice & .867 & .860 & $\mathbf{-.008}$ (9)  & $\mathbf{5.1\text{e-}4}$\\
 & CSF   & .898 & .891 & $-.007$ (10) & $4.7\text{e-}3$\\
 & GM    & .838 & .837 & $-.001$ (18) & .28\\
 & WM    & .867 & .851 & $\mathbf{-.016}$ (10) & $\mathbf{2.0\text{e-}3}$\\
\midrule
\multirow{4}{*}{\rotatebox{90}{Swin}}
 & mDice & .874 & .878 & $+.004$ (22) & .32\\
 & CSF   & .907 & .903 & $-.004$ (16) & .27\\
 & GM    & .849 & .852 & $+.003$ (23) & .16\\
 & WM    & .867 & .878 & $+.012$ (24) & .16\\
\midrule
\multirow{4}{*}{\rotatebox{90}{WaveMix}}
 & mDice & .858 & .872 & $\mathbf{+.014}$ (30) & $\mathbf{3.5\text{e-}5}$\\
 & CSF   & .886 & .904 & $\mathbf{+.018}$ (36) & $\mathbf{7.8\text{e-}7}$\\
 & GM    & .834 & .847 & $\mathbf{+.013}$ (36) & $\mathbf{3.2\text{e-}6}$\\
 & WM    & .854 & .865 & $+.011$ (26) & .10\\
\bottomrule
\end{tabular}
\end{table}

\subsection{Effect of the enhancer variant}
\begin{figure*}[t]
\centering
\includegraphics[width=0.8\textwidth]{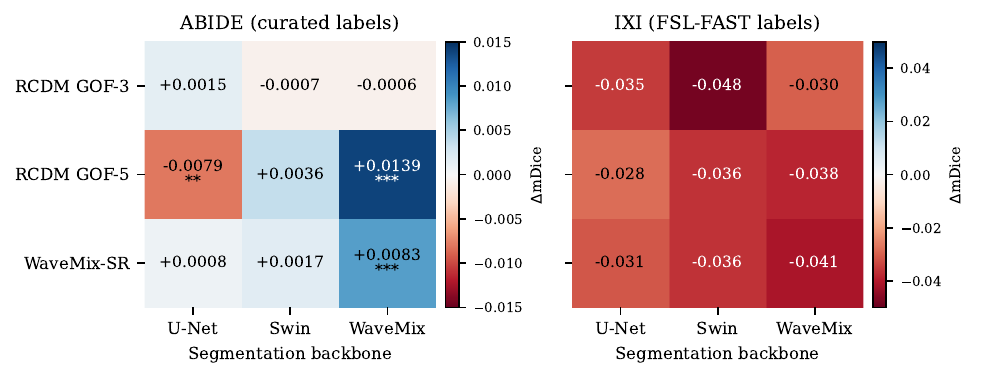}
\caption{Change in mean Dice for every enhancer variant $\times$ segmentation
backbone. Left: ABIDE (FreeSurfer labels); stars mark significant paired
Wilcoxon tests. Right: IXI (FSL-FAST labels): all pairings lose.}\label{fig:grid}
\end{figure*}

Fig.~\ref{fig:grid} (left) crosses all three enhancer variants with the three
backbones. The wavelet-segmenter column contains the two largest gains (GOF-5:
$+0.0139$; WaveMix-SR: $+0.0083$, $p=7\times10^{-4}$), and the only
significant loss is GOF-5 with the U-Net. The GOF\,=\,3 enhancer changes no
backbone by more than 0.0015, whereas the architecturally almost identical
GOF\,=\,5 enhancer changes two backbones significantly in opposite directions.

\begin{figure*}[t]
\centering
\includegraphics[width=0.9\textwidth]{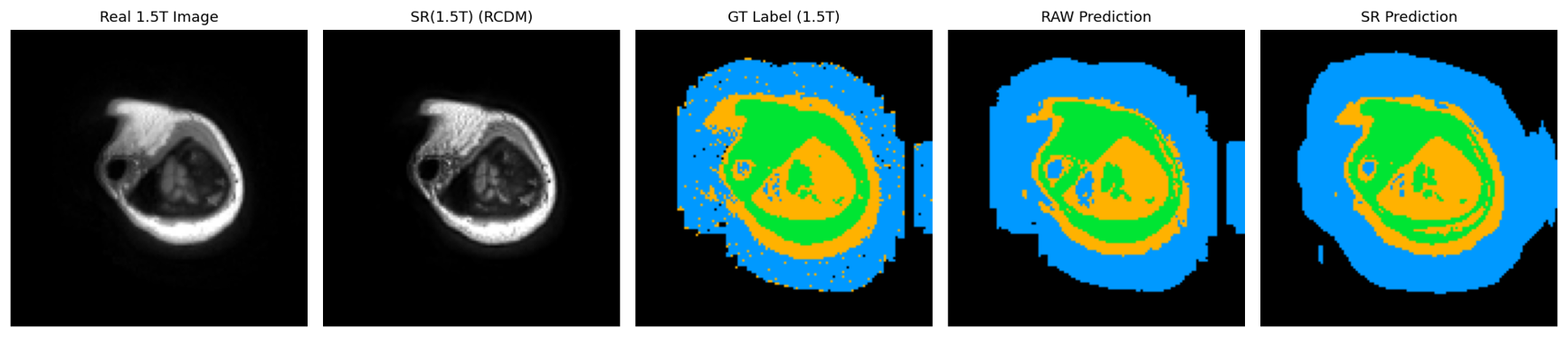}
\caption{ABIDE example (axial). From left: real 1.5\,T, enhanced, label, RAW
prediction, SR prediction (blue CSF, green GM, orange WM).
}\label{fig:abidequal}
\end{figure*}

Fig.~\ref{fig:abidequal} shows a representative ABIDE slice.

\subsection{Segmentation on IXI and label quality}
On IXI every pairing loses mean Dice, by 0.028 to 0.048
(Fig.~\ref{fig:grid}, right). For the GOF\,=\,5 enhancer the class-wise changes
are nearly identical across backbones: CSF falls by about 0.10, GM changes by
less than 0.012 in either direction, and WM is unchanged. A loss that is
confined to one class and is the same for three architectures points to the
data rather than to any model.

\begin{figure*}[t]
\centering
\includegraphics[width=0.9\textwidth]{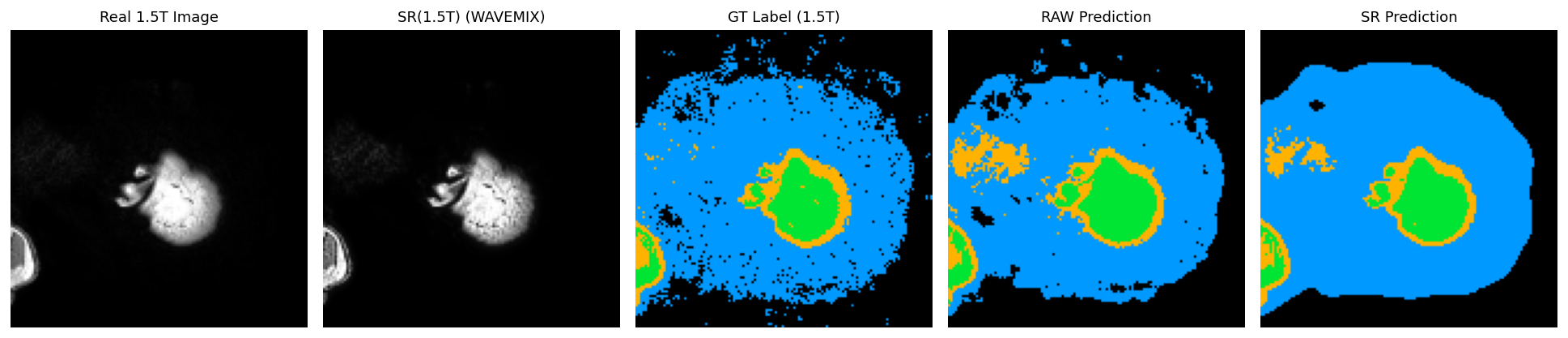}
\caption{IXI example. From left: real 1.5\,T, enhanced, FSL-FAST label, RAW
prediction, SR prediction. Isolated CSF voxels (blue) inside tissue in the
label are not visible in the image.}\label{fig:ixi}
\end{figure*}

\begin{figure*}[t]
\centering
\includegraphics[width=0.9\textwidth]{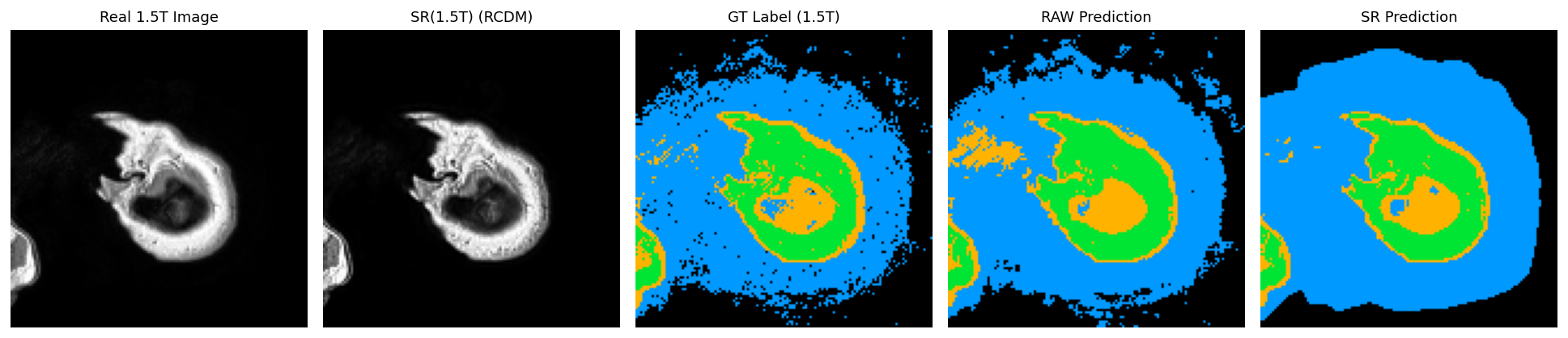}
\caption{A second IXI slice with the same layout; the SR prediction is
smoother than the label.}\label{fig:ixi2}
\end{figure*}

Figs.~\ref{fig:ixi} and~\ref{fig:ixi2} show the reason. The FSL-FAST labels
contain CSF voxels scattered through the brain parenchyma with no counterpart
in the image; the \textsc{raw} predictions reproduce some of this speckle,
and the \textsc{sr} predictions, trained on smoother images, reproduce less of
it. Dice against such labels rewards reproducing label noise. IXI therefore
tells us about the labels, not about the enhancer. We also note secondary
factors that may contribute: scanner contrast outside the training
distribution of the enhancer, the absence of site-specific intensity
harmonisation before enhancement, and the $128\times128$ segmentation
resolution, which discards part of any boundary improvement.


\section{Discussion}\label{sec:discussion}
\textbf{Fidelity does not predict segmentation benefit.} The enhancer
raises fidelity on every validation epoch, yet its effect on segmentation
ranges from significantly positive to significantly negative depending on the
segmenter, and is nil for the GOF\,=\,3 variant. A fidelity gain alone would
therefore have predicted none of the downstream outcomes.

\textbf{Why does the effect depend on the segmenter?} The wavelet segmenter,
whose token mixing operates on the same Haar sub-bands that the enhancer is
conditioned on, gains most, and its gains are in CSF and GM, the classes most
affected by partial-volume blurring. One reading is that enhancement
restores exactly the band-limited boundary information that a wavelet mixer
uses. A second reading fits the data as well: the wavelet segmenter has the
lowest \textsc{raw} score, so enhancement may mainly let a weaker model catch
up with the others, whose scores it approximately reaches. The U-Net result is
harder to explain by either reading; one possibility is that sharper
boundaries in the enhanced images no longer coincide with the FreeSurfer
boundaries, which were defined on the original 1.5\,T geometry, and that a
local convolutional model fits those offsets more closely. Our experiment does
not separate these mechanisms. Doing so would require more backbones per
family, capacity-matched controls, and labels defined on 3\,T scans of the
same subjects.

\textbf{Label quality is part of the evaluation.} The IXI result shows that an
automated label source can turn a regularising preprocessing step into an
apparent failure. A study that used only IXI would have concluded that
enhancement harms segmentation; one that used only ABIDE and the wavelet
segmenter would have concluded the opposite. Reporting several backbones and
checking labels visually are cheap safeguards against both errors.

\textbf{Limitations.} (i)~Segmentation is slice-based at $128\times128$; a
3D evaluation at native resolution is needed. (ii)~Within each experiment each
segmenter is trained once (seed 42). Across our three experiments, which share
the split, the independently retrained \textsc{raw} U-Net and Swin-UNet
differ by up to 0.006 mean Dice, comparable to their SR effects but much
smaller than the wavelet-segmenter gain; repeated seeds would tighten these
estimates. (iii)~Fidelity is
measured on synthetic pairs only; no paired same-subject 1.5\,T/3\,T scans were
available. (iv)~ABIDE labels are FreeSurfer outputs rather than manual
annotations. (v)~Both cohorts are research datasets; clinical 1.5\,T data with
pathology were not studied.

\section{Conclusion}\label{sec:conclusion}
We presented a physics-guided, unpaired training pipeline for a small 1.5\,T
brain MRI enhancer and evaluated it by what it does to tissue segmentation.
Enhancement significantly improved a wavelet token-mixing segmenter,
significantly degraded a U-Net and left a Swin-UNet unchanged on the same
subjects and labels, and appeared harmful on a dataset with automated labels
for reasons traceable to the labels. Enhancement methods for low-field MRI
should be reported with several downstream models, with paired statistics,
and against labels whose quality has been checked.


\section*{Code and data availability}
ABIDE and IXI are publicly available from their providers. Code, trained
models, data splits and per-subject results will be released on GitHub.

\bibliographystyle{plainnat}
\bibliography{refs}

@article{jeevan2023wavemix,
  title={WaveMix: A resource-efficient neural network for image analysis},
  author={Jeevan, Pranav and Viswanathan, Kavitha and Sethi, Amit and others},
  journal={arXiv preprint arXiv:2205.14375}, year={2023}
}

@inproceedings{zhu2019dcnv2,
  title={Deformable ConvNets v2: More deformable, better results},
  author={Zhu, Xizhou and Hu, Han and Lin, Stephen and Dai, Jifeng},
  booktitle={IEEE/CVF Conference on Computer Vision and Pattern Recognition (CVPR)}, pages={9308--9316}, year={2019}
}

@inproceedings{viswanathan2025lowresource,
  title={Low-Resource Video Super-Resolution with Memory, Wavelets, and Deformable Convolutions},
  author={Viswanathan, Kavitha and Sethi, Amit and Pathak, Shashwat and Bharambe, Piyush and Choudhary, Harsh},
  booktitle={IEEE/CVF CVPR Workshops, Women in Computer Vision (WiCV)}, year={2025}
}

@article{martino2014abide,
  title={The autism brain imaging data exchange: Towards a large-scale evaluation of the intrinsic brain architecture in autism},
  author={Di Martino, Adriana and Yan, Chao-Gan and Li, Qingyang and Denio, Erin and Castellanos, Francisco X and others},
  journal={Molecular Psychiatry}, volume={19}, number={6}, pages={659--667}, year={2014}
}

@misc{ixi2025,
  title={{IXI} Dataset: T1, T2 and PD-weighted brain {MRI}},
  author={{Imperial College London Biomedical Image Analysis Group}},
  howpublished={\url{https://brain-development.org/ixi-dataset/}}, year={2025}
}

@misc{ixi,
  title        = {{IXI} Dataset --- Brain Development},
  howpublished = {\url{https://brain-development.org/ixi-dataset/}},
  year         = {2025}
}

@inproceedings{ronneberger2015unet,
  author    = {Olaf Ronneberger and Philipp Fischer and Thomas Brox},
  title     = {{U-Net}: Convolutional Networks for Biomedical Image Segmentation},
  booktitle = {MICCAI},
  year      = {2015}
}

@inproceedings{cao2022swinunet,
  author    = {Hu Cao and Yueyue Wang and Joy Chen and Dongsheng Jiang and Xiaopeng Zhang and Qi Tian and Manning Wang},
  title     = {{Swin-Unet}: Unet-like Pure Transformer for Medical Image Segmentation},
  booktitle = {ECCV Workshops},
  year      = {2022}
}

@article{zhang2001fast,
  author  = {Yongyue Zhang and Michael Brady and Stephen Smith},
  title   = {Segmentation of Brain {MR} Images Through a Hidden {Markov} Random Field Model and the Expectation-Maximization Algorithm},
  journal = {IEEE Transactions on Medical Imaging},
  volume  = {20},
  number  = {1},
  pages   = {45--57},
  year    = {2001}
}

@inproceedings{zhu2017cyclegan,
  author    = {Jun-Yan Zhu and Taesung Park and Phillip Isola and Alexei A. Efros},
  title     = {Unpaired Image-to-Image Translation Using Cycle-Consistent Adversarial Networks},
  booktitle = {ICCV},
  year      = {2017}
}

@inproceedings{wang2018patchgan,
  author    = {Ting-Chun Wang and Ming-Yu Liu and Jun-Yan Zhu and Andrew Tao and Jan Kautz and Bryan Catanzaro},
  title     = {High-Resolution Image Synthesis and Semantic Manipulation with Conditional {GANs}},
  booktitle = {CVPR},
  year      = {2018}
}

@article{ashesh2024microssim,
  title   = {{MicroSSIM}: Improved Structural Similarity for Comparing Microscopy Data},
  author  = {Ashesh, Ashesh and Krull, Alexander and Jug, Florian},
  journal = {arXiv preprint arXiv:2408.08747},
  year    = {2024}
}

@article{fischl2012freesurfer, author={Fischl, Bruce}, title={{FreeSurfer}}, journal={NeuroImage}, volume={62}, number={2}, pages={774--781}, year={2012}}

@inproceedings{mao2017lsgan, author={Mao, Xudong and Li, Qing and Xie, Haoran and Lau, Raymond Y. K. and Wang, Zhen and Smolley, Stephen Paul}, title={Least Squares Generative Adversarial Networks}, booktitle={IEEE International Conference on Computer Vision (ICCV)}, pages={2794--2802}, year={2017}}

@article{wilcoxon1945, author={Wilcoxon, Frank}, title={Individual comparisons by ranking methods}, journal={Biometrics Bulletin}, volume={1}, number={6}, pages={80--83}, year={1945}}

@inproceedings{chen2018mdcsrn, author={Chen, Yuhua and Shi, Feng and Christodoulou, Anthony G. and Xie, Yibin and Zhou, Zhengwei and Li, Debiao}, title={Efficient and Accurate {MRI} Super-Resolution Using a Generative Adversarial Network and {3D} Multi-level Densely Connected Network}, booktitle={Medical Image Computing and Computer Assisted Intervention (MICCAI)}, series={LNCS}, volume={11070}, pages={91--99}, year={2018}}

@inproceedings{pham2017brainsr, author={Pham, Chi-Hieu and Ducournau, Aur{\'e}lien and Fablet, Ronan and Rousseau, Fran{\c{c}}ois}, title={Brain {MRI} super-resolution using deep {3D} convolutional networks}, booktitle={IEEE International Symposium on Biomedical Imaging (ISBI)}, pages={197--200}, year={2017}}

@article{zhao2021smore, author={Zhao, Can and Dewey, Blake E. and Pham, Dzung L. and Calabresi, Peter A. and Reich, Daniel S. and Prince, Jerry L.}, title={{SMORE}: A Self-supervised Anti-aliasing and Super-resolution Algorithm for {MRI} Using Deep Learning}, journal={IEEE Transactions on Medical Imaging}, volume={40}, number={3}, pages={805--817}, year={2021}}

@article{iglesias2023synthsr, author={Iglesias, Juan Eugenio and Billot, Benjamin and Balbastre, Ya{\"e}l and Magdamo, Colin and Arnold, Steven E. and Das, Sudeshna and Edlow, Brian L. and Alexander, Daniel C. and Golland, Polina and Fischl, Bruce}, title={{SynthSR}: A public {AI} tool to turn heterogeneous clinical brain scans into high-resolution {T1}-weighted images for {3D} morphometry}, journal={Science Advances}, volume={9}, number={5}, pages={eadd3607}, year={2023}}

@article{billot2023synthseg, author={Billot, Benjamin and Greve, Douglas N. and Puonti, Oula and Thielscher, Axel and Van Leemput, Koen and Fischl, Bruce and Dalca, Adrian V. and Iglesias, Juan Eugenio}, title={{SynthSeg}: Segmentation of brain {MRI} scans of any contrast and resolution without retraining}, journal={Medical Image Analysis}, volume={86}, pages={102789}, year={2023}}

@article{lin2023lowfieldiqt, author={Lin, Hongxiang and Figini, Matteo and D'Arco, Felice and Ogbole, Godwin and Tanno, Ryutaro and Blumberg, Stefano B. and Ronan, Lisa and Brown, Biobele J. and Carmichael, David W. and Lagunju, Ikeoluwa and Cross, Judith Helen and Fernandez-Reyes, Delmiro and Alexander, Daniel C.}, title={Low-field magnetic resonance image enhancement via stochastic image quality transfer}, journal={Medical Image Analysis}, volume={87}, pages={102807}, year={2023}}

@article{alexander2017iqt, author={Alexander, Daniel C. and Zikic, Darko and Ghosh, Aurobrata and Tanno, Ryutaro and Wottschel, Viktor and Zhang, Jiaying and Kaden, Enrico and Dyrby, Tim B. and Sotiropoulos, Stamatios N. and Zhang, Hui and Criminisi, Antonio}, title={Image quality transfer and applications in diffusion {MRI}}, journal={NeuroImage}, volume={152}, pages={283--298}, year={2017}}

@article{maierhein2024metrics, author={Maier-Hein, Lena and Reinke, Annika and Godau, Patrick and others}, title={Metrics reloaded: recommendations for image analysis validation}, journal={Nature Methods}, volume={21}, number={2}, pages={195--212}, year={2024}}

@article{gudbjartsson1995rician, author={Gudbjartsson, H{\'a}kon and Patz, Samuel}, title={The {R}ician distribution of noisy {MRI} data}, journal={Magnetic Resonance in Medicine}, volume={34}, number={6}, pages={910--914}, year={1995}}
\end{document}